\documentclass[conference]{IEEEtran}
\IEEEoverridecommandlockouts
\usepackage{cite}
\usepackage{amsmath,amssymb,amsfonts}
\usepackage{algorithmic}
\usepackage{subcaption}
\usepackage{graphicx}
\usepackage{comment}
\usepackage{colortbl}
\usepackage{textcomp}
\usepackage{listings}
\usepackage{xcolor}
\usepackage{hyperref}

\def\BibTeX{{\rm B\kern-.05em{\sc i\kern-.025em b}\kern-.08em
    T\kern-.1667em\lower.7ex\hbox{E}\kern-.125emX}}

\newcommand{\mona}[1]{\textcolor{blue}{\textit{\textbf{Mona}: #1}}}

\begin{document}\vspace{-35pt}
\title{Test Case Recommendations with Distributed Representation of Code Syntactic Features
}\vspace{-35pt}
\author{\IEEEauthorblockN{Mosab Rezaei, Hamed Alhoori, Mona Rahimi}
\IEEEauthorblockA{\textit{Dept. of Computer Science} \\
\textit{Northern Illinois University}\\
%DeKalb, IL \\
mosab.rezaei, alhoori, mrahimi1@niu.edu}
%\and
%\IEEEauthorblockN{2\textsuperscript{nd} Hamed Alhoori}
%\IEEEauthorblockA{\textit{Dept. of Computer Science} \\
%\textit{Northern Illinois University}\\
%Dekalb, US \\
%alhoori@niu.edu}
%\and
%\IEEEauthorblockN{3\textsuperscript{rd} Mona Rahimi}
%\IEEEauthorblockA{\textit{Dept. of Computer Science} \\
%\textit{Northern Illinois University}\\
%Dekalb, US \\
%mrahimi1@niu.edu}
}
\maketitle
\vspace{-35pt}
\begin{abstract}
Frequent modifications of unit test cases are inevitable due to software's continuous underlying changes in source code, design, and requirements. Since manually maintaining software test suites is tedious, timely, and costly, automating the process of generation and maintenance of test units will significantly impact the effectiveness and efficiency of software testing processes. 

To this end, we propose an automated approach which exploits both structural and semantic properties of source code methods and test cases to recommend the most relevant and useful unit tests to the developers. 
The proposed approach initially trains a neural network to transform method-level source code, as well as unit tests, into distributed representations (embedded vectors) while preserving the importance of the structure in the code. Retrieving the semantic and structural properties of a given method, the approach computes cosine similarity between the method's embedding and the previously-embedded training instances. Further, according to the similarity scores between the embedding vectors, the model identifies the closest methods of embedding and the associated unit tests as the most similar recommendations.

The results on the Methods2Test dataset showed that, while there is no guarantee to have similar relevant test cases for the group of similar methods, the proposed approach extracts the most similar existing test cases for a given method in the dataset, and evaluations show that recommended test cases decrease the developers' effort to generating expected test cases. 
\end{abstract}


\section{Introduction}
\label{Introduction}
A test case is a set of conditions designed to determine whether a feature or functionality of the software is working as expected. Each unit of a source code, such as a method, is tested by an individual test case known as a unit test. Unit tests are executed to compare the actual output of a relevant method with a value expected by the developers. 

Manually keeping updated test cases is tedious, time-consuming, and costly. Hence, several works have attempted to automate the process, like generating new test cases based on their older versions \cite{mirzaaghaei2012supporting}. Among them, 
there are only a few, which provide developers with functioning tools to generate test cases, such as EvoSuite \cite{fraser2011evosuite}, and Randoop \cite{pacheco2007randoop}. EvoSuite uses a search-based approach, while Randoop is based on a random feedback-oriented technique. While these tools are fairly known to the developers, multiple independent researchers reported several issues with their generated test cases, such as the program coverage or meeting the developers' expectations \cite{fraser2015does, shamshiri2015automatically}.

More recently, the convolutional neural networks (CNN) and embedding techniques received special attention for the analysis of code-base artifacts. For instance, the func2vec model \cite{defreez2018path} maps each code function to a vector in a multi-dimensional space in such a way that functions with similar functionality are placed closer to each other.

Neural networks are as well applied for the creation of semantic relationships between a method in the source code and a descriptive method name in natural language \cite{allamanis2015suggesting}. A recent model, namely TreeCaps \cite{bui2021treecaps}, uses a Tree-based convolutional neural network embedding to learn the source code better, achieving higher performance in both code functionality classification and method name prediction tasks. Another successful model in code fragment embedding, adopted in this research, is called code2vec \cite{alon2019code2vec}. The model generates an embedded vector for a given code snippet so as to predict a descriptive title as the method’s name. The code2vec model is reflected and adopted in a few other studies like \cite{allamanis2019adverse} to evaluate the effect of code duplication.

In this work, given the success stories of the code embedding approaches, we attempted to leverage an embedding technique to develop a test case recommendation application, which considers the structural and semantic similarities between the methods as a metric for its recommendations. 

The leverage of the similarity between the embedded vectors as a recommendation metric is common in the literature. For instance, cosine similarity of the generated vectors by word2vec \cite{mikolov2013distributed}, is measured in a fashion recommender system~\cite{kavitha2020fashion}.  In a similar work \cite{ramyasree2021code}, the similarity between generated embedded vectors by code2vec has been exploited to recommend a code snippet for a given query. However, to the best of our knowledge, this is the first attempt to use the code distributed representation and their similarities to recommend test cases.

The proposed approach is inspired by a model known as code2vec, which considers the code's Abstract Syntax Tree (AST) to extract structural and semantic features of the code. The model generates a single fixed-length vector (i.e. embedded vector) from each code snippet and maps the embeddings to a multi-dimensional space such that the similar methods, structurally and semantically, are placed closer to each other in the space. The model is originally proposed for predicting the name of the methods.

To measure the similarity between the code snippets, each method in the source code, as well as each unit test in the associated test suite, were converted to a single fixed-length code vector based on the code's structural and semantic properties, to be later mapped in a  high-dimensional space. 
The distribution of the embedded methods and test cases within a single space allows for measuring the similarities between method pairs, test pairs, as well as the method and test pairs, searching the space for potentially relevant test cases to methods that are mostly similar. Taking into consideration both semantic and structural similarities, we aim to recommend test cases that satisfy developers' needs, looking for the best-matched test cases to their developed functionalities.

To evaluate the proposed approach, we calculate the similarity between recommended test cases and expected test cases for all methods in the dataset. 
There are two primary research questions, which this paper is probing to answer: 

\begin{itemize}

\item{$RQ_1 (Functionality):$} Given the structural and semantic code embeddings, a continuous vector representation of code snippets, can we train a neural network to recommend a relevant test case for a given method?

This research question evaluates two alternative approaches for the purpose of recommending the most relevant test case for each given method. The first approach measures the similarity between the given method with previously-seen methods, while the second approach uses the similarity between the given method and pre-embedded test cases. Each approach, therefore, relies on the assumption that a meaningful relationship exists between the distributed representation of either similar methods or similar methods and unit tests.

\item {$RQ_2 (Usability):$} How useful are the proposed test cases in reducing developers' efforts in test suit maintenance?

This research question measures the usefulness of the recommended unit tests in the software development process. In order to answer this question, Levenshtein distance has been used to calculate the distances between the strings of recommended test cases and expected test cases. It has been used to demonstrate how much the syntax of test cases, which developers need, are similar to each other.
%Regardless of the answer to the first research question, we are going to know how much the recommended test cases are helpful and applicable for developers. Based on our main goal, which was proposing a test case recommender, in this paper, we are going to know how much the recommended test cases can decrease the developers' effort while creating and running new test cases for their methods.
\end{itemize}

To answer the research questions, we developed two variants of an application which differ in their assumption for prioritizing the recommendations. The implementation and the evaluation results of the two approaches further specify whether similar functions, such as two sort functions (bubble sort and quick sort), require semantically and structurally similar unit tests or various functions, such as sort and copy, tend to require unit tests associated with \textit{arrays} or \textit{matrix}, depending on the structure of the data, which is being used in the function. Finding an answer to this question is essential since, in the given example, the answer identifies that recommending unit tests relevant to a method functions (sorting) are more useful or the unit tests relevant to the structure which is being tested which test the structure under test (a matrix structure).

The rest of this paper is organized as follows. Section \ref{related} is dedicated to the related works. Section \ref{Vectors} provides the essential required information for this paper. In section \ref{Approach} proposed approach has been explained. After that, in section \ref{evaluation} the evaluation steps and their results have been reported. Finally, in section \ref{Validity} and \ref{conclusion} threats to validity and conclusion are explained, respectively. 
Due to the limitation of the space, the artifacts of this work are accessible on our online repository\footnote{https://github.com/mosabrezaei/Test-Case-Recommendation}.

\section{Related Work}
\label{related}
%The processes of executing programs, or part of programs, with the intention of finding bugs and defects, are categorized under the topic of software testing \cite{myers2011art}. 
The importance of developing new techniques to facilitate the process of software testing in the industry and in research is known in the software engineering community. Despite significant contributions from both academia and industrial environments on various aspects of software, testing \cite{barr2014oracle}, yet several concerns exist which require to be addressed\cite{bluemke2021software}. One of the most challenging areas in software testing is known to be the automation of generating accurate and useful test cases.

There are a wide variety of approaches developed and discussed in the literature to automate the generation and recommendation of test cases.  
One commonly-used method, often adopted for the purpose of evolving test cases to a newer version of the software, is to extract and leverage the necessary information from the software's older test suites. For instance, in  \cite{mirzaaghaei2012supporting}, the authors utilized the information of test cases present in an older version of programs to repair and generate new test cases for the recent version of the programs.

With the increasing success of neural networks, researchers in several areas attempted to address various problems with the adoption of the artificially-intelligent models. For instance, one known model in this area is word2vec \cite{mikolov2013distributed}, which enabled researchers to improve the  natural language processing (NLP) tasks. 

Furthermore, due to the success of the deep neural networks, researchers have recently started applying these models to answer a variety of research questions around the source code. For instance, one of them is CodeBERT \cite{feng2020codebert}. CodeBERT is a bimodal pre-trained model for natural language and programming language. It can be adopted in several downstream tasks, such as code captioning. Another example is AthenaTest \cite{tufano2020unit}. AthenaTest is designed to generate test cases for given methods in Java.

To apply neural network-based techniques to the source code, it is necessary to initially represent the source code fragments in the form of numerical vectors. In this regard, the embedding techniques tend to transfer the source code into a new space with lower dimensions than the original space, in which each code fragment is represented with a continuous vector. To this end, several embedding techniques are proposed, such as the embedding of functions and binary code\cite{chen2019literature}. Furthermore, in \cite{efstathiou2019semantic}, the authors published the applications and limitations of each method in this area by applying the models on the source code of different programming languages.

Additionally, there are models that are trained in particular only to embed source code, such as CuBERT \cite{kanade2020learning} or CODEnn \cite{gu2018deep}. Moreover, in \cite{defreez2018path} the func2vec model has been proposed, which distributes the methods through the space in a way that the embedded vector of the functions with the same functional similarity is closer to each other. In \cite{allamanis2015suggesting} authors created semantic relationships between the name of the methods. They achieve to generate new method names never seen in the training data. One of the interesting models in the code embedding area is the code2vec, which leverages the semantics of the primary paths of a code fragment to generate the embedding \cite{alon2019code2vec}. Authors have shown that code2vec can generate accurate vectors for each snippet code by applying it in a label prediction task.
\section{Continuous Distributed Vectors}
\label{Vectors}
Methods for learning distributed representations produce low-dimensional vector representations of elements with several modalities, such as language, vision, and code. As such, the meaning (semantics) of an element from different modalities is distributed across multiple vector components, such that semantically similar objects are mapped to close vectors. 

\subsection{Numerical Continuous Representation}
In natural language processing, several models are developed to generate dense continuous vector representations of textual data in a vector space, such as word2vec \cite{mikolov2013distributed}. A numerical continuous representation of words provides the opportunity to develop solutions based on machine learning and neural networks. These models transform each word to a continuous vector in a space with relatively small dimensions.

While the same NLP models are applicable to the code-based artifacts (\textit{e.g.,} method and unit tests) as well, once applied to the code, the structural properties of the source code are ignored, and only the semantic information of the words is considered~\cite{chen2019literature}. This is a primary limitation of applying NLP models to the code-based artifacts since the structure of the code is, in fact, informative about the contents of the code. For instance, two code snippets with similar semantics but different structures may provide different functionalities.
Several models in the literature are developed in particular for the transformation of code fragments into a single vector of embedding \cite{chen2019literature}.

\subsection{The Code2vec Model}

An Abstract Syntax Tree (AST) is a tree representation of a code snippet,  reflecting the structure of the source code. A recent work, proposed code2vec, using the AST structure of the code to identify the paths in the graph which are more informative of the method's functionality~\cite{alon2019code2vec}. 

\subsubsection{\textbf{Abstract Syntax Tree}}
For each code snippet \begin{math}C\end{math}, there is an AST which express as \begin{math}\langle N,T,X,s,\delta,\phi\rangle\end{math}. At the expressed tuple, \begin{math}N\end{math}, \begin{math}T\end{math}, and \begin{math}X\end{math} are the sets of the non-terminal nodes (non-leaf), terminal nodes (leaf), and values respectively. The \begin{math}s\end{math} is the root node. Also, the \begin{math}\delta\end{math} and \begin{math}\phi\end{math} are the mapping functions. the \begin{math}\delta\end{math} maps a non-terminal node to a list of its children, and the \begin{math}\phi\end{math} maps terminal node to an associated value respectively.

\subsubsection{\textbf{AST-path}}
The \begin{math}p\end{math} is an AST-path which is simply a path between two leaves in AST. In other words, a path with the length of \begin{math}k\end{math} is a sequence of the form: \begin{math}p = n_{1}d_{1}...n_{k}d_{k}n_{k+1}\end{math}, where both \begin{math}n_{1}\end{math} and \begin{math}n_{k+1}\end{math} are terminal nodes. All the other \begin{math}n\end{math} are non-terminal nodes. Also, \begin{math}d\end{math} are the direction of the moving through the tree that can be up or down. Given the sequence of \begin{math}p\end{math}, \begin{math}start(p)\end{math} and \begin{math}end(p)\end{math} refer to \begin{math}n_1\end{math} and \begin{math}n_{k+1}\end{math} respectively. 

\subsubsection{\textbf{Path-context}}
For any arbitrary AST-path \begin{math}p\end{math}, there is a set of Path-contexts \begin{math}b\end{math} known as a bag of Path-contexts. Each Path-context \begin{math}b_i\end{math} express as triple \begin{math}<x_s,p,x_t>_i\end{math}. \begin{math}x_s\end{math} is the value of \begin{math}n_1\end{math} and \begin{math}x_t\end{math} is the value of \begin{math}n_{k+1}\end{math}. Therefore, Path-context contains two real values from the code snippet and one syntactic path that connects these values together based on the extracted AST.

\subsubsection{\textbf{Context Vector}}
The embedded vector for the single path-context \begin{math}b_i = <x_s,p_j,x_t>\end{math} known as context vector \begin{math}C_i\end{math}. As shown below equation, the context vector is created by concatenating three smaller embedded vectors that refer to the triple \begin{math}b_i\end{math} components.
%\vspace{-5pt}
\begin{equation}\footnotesize C_i=E(b_i)=[value\_vocab_s;path\_vocab_j;value\_vocab_t]\end{equation}

During training, code2vec learns the following embedded matrices: paths (\begin{math}path\_vocab\end{math}), values(\begin{math}value\_vocab\end{math}), labels, fully connected layer, and attention vector \begin{math}a\end{math}. The \begin{math}Embedding (b_i)\end{math} extracts embedded vectors from the embedded matrices, \begin{math}path\_vocab\end{math} and \begin{math}value\_vocab\end{math}.

\subsubsection{\textbf{Combined Context Vector}}
Each context vector \begin{math}C_i\end{math}, which is created by concatenating three different embedded vectors, will pass through a fully connected layer to generate Combined Context Vector \begin{math}\tilde{C}_i\end{math}.

\subsubsection{\textbf{Code Vector}}
Eventually, to generate code vector V, which is the final version of embedding snippet code C, there is a fully connected layer that exploits the attention mechanism.
To aim this goal, all generated combined context vectors \begin{math}\{\tilde{C}_1,...,\tilde{C}_n\}\end{math} will be aggregated into a single vector. Finally, the code vector V will be generated by accumulating of weighted combined context vector.

It needs to be mentioned that there is one more step after generating code vector V, which is generating a prediction vector. The related equation for this step has mentioned in \cite{alon2019code2vec, coimbra2021using}. After fine-tuning the model, the code vector V for code snippet C has been extracted \begin{math}V = Embedding(C)\end{math} to be used in the proposed approaches. Code2vec is pre-trained based on 10,072 Java GitHub repositories, including 13,376,807 samples with respect to both fair measures of code similarity. The model is shown to outperform other models in several tasks, such as in detecting the C source code vulnerability.

\begin{figure}[ht]
\vspace{-5pt}
\centerline{\includegraphics[width=.95\linewidth]{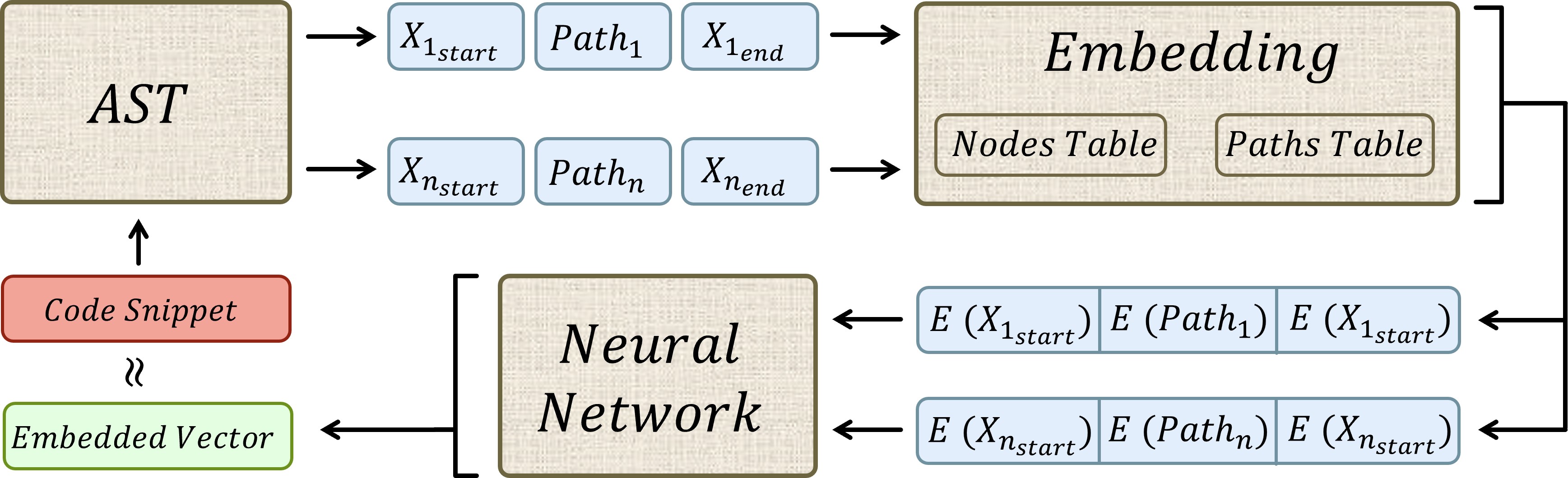}}
%\vspace{-5pt}
\caption{An overview of code2vec Architecture.}
\label{code2vec}
\vspace{-10pt}
\end{figure}

Figure ~\ref{code2vec} represents an overview of the code2vec architecture. As shown, the code snippets are initially transformed to AST representations. As discussed in section \ref{Vectors}, the context of the primary paths, reflecting the structural information, will be extracted from the AST representation of the code snippets. Each extracted path is then embedded by a shallow neural network, which augments the embedding with the semantic information of each path. The neural network's output is, therefore, a single fixed-length vector containing semantic and structural information about the code.

\begin{figure*}[t]
\vspace{-10pt}
\centerline{\includegraphics[scale=.5]{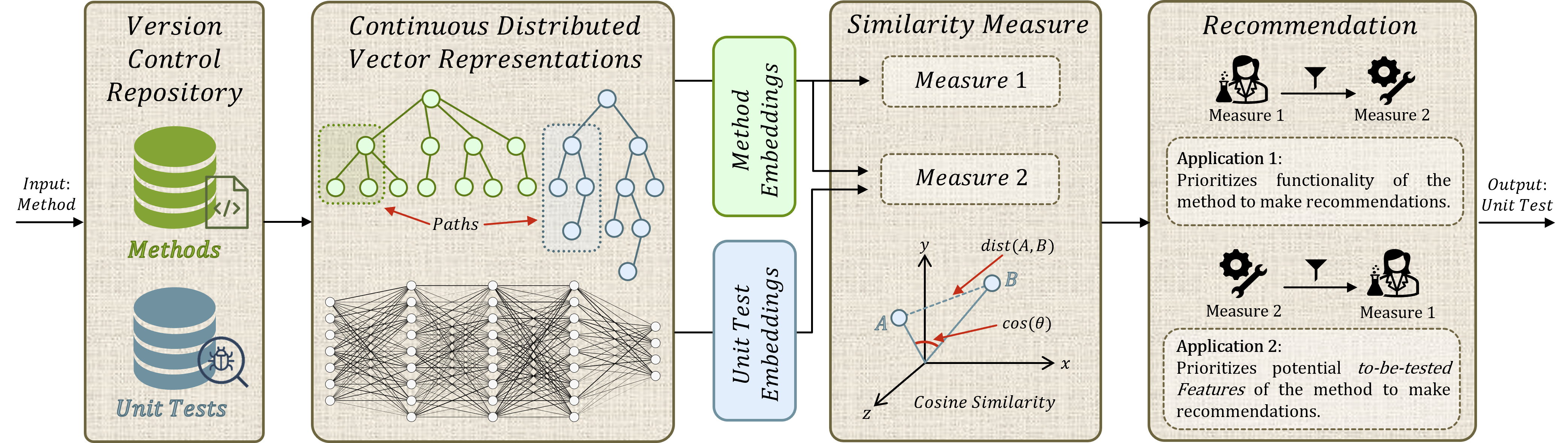}}
%\captionsetup{justification=centering}
\caption{A High-level overview of the proposed Approach.}
\label{overview}
\vspace{-10pt}
\end{figure*}
\section{Proposed Approach}
\label{Approach}
\definecolor{Gray}{gray}{0.85}
This section discusses each of the two approaches which we implemented to answer our research questions. Both of the proposed approaches initially transform the dataset code snippets into embedding to calculate the distance between the embedded vectors as a similarity metric.The overview of the proposed approaches has been presented in Figure \ref{overview}. 

\subsection{Representing Code Snippets}

The code structure is crucial in determining the code contexts to accurately measure their similarity and dissimilarity. Hence we selected the cod2vec \cite{alon2019code2vec} to represent code snippets in continuous vectors by considering both structural and semantic information.

\subsubsection{Pre-Processing the Dataset} 
To build a model for the purpose of test case recommendation, we initially retrained the model with a set of correspondent source and test code fragments. For this, we used the Methods2Test dataset, containing 577,611 samples \cite{tufano2020unit}. Below, a sample of the dataset has been shown:

%\vspace{1mm}

\setlength{\parindent}{0ex}

\begin{flushleft}
\textbf{\small Methods:}

~~~~\noindent {\fontfamily{pcr}\footnotesize\selectfont public boolean isEmpty() \{ return 0 == size; \}}
\end{flushleft} 

\begin{flushleft}
\textbf{\small Test Cases:}

~~~~\noindent {\fontfamily{pcr}\footnotesize\selectfont @Test public void shouldReportEmpty() \{\\~~~assertThat(map.isEmpty(), is(true)); ~\}}
\end{flushleft} 

\setlength{\parindent}{2ex}

A generated embedded vector contains 384 elements representing the method (test case) as a point in a 384-dimensional space.
We noticed that the original code2vec model generates two dissimilar vectors for multiple individual methods, such as nested methods. In addition, several syntax errors in the dataset,  such as methods ended with empty curly brackets, made the model fail in generating the distributed vectors for the entire instances. The process of cleaning the dataset for the model was one of the tedious parts of the pre-processing activities. Finally, 4,500 cleaned samples were extracted from the Methods2Test dataset.

\iffalse
For instance, the Java method below converts Mile to Kilometer:

\vspace{3mm}

\lstdefinestyle{mystyle}{
tabsize=3, 
breaklines=true,
showstringspaces=false,
numbersep=8pt}
\lstset{style=mystyle}

\begin{lstlisting}[language=Java, basicstyle=\small, basicstyle=\footnotesize\ttfamily]
public double MileToKm(double mile){
	if (mile < 0) 
	   throw new IllegalArgumentException("Argument must be positive.");
	double km = mile * 1.609344;
	return km;} 
\end{lstlisting}

%\vspace{3mm}

To fully test the functionality of the given method ``MileToKm'', a large number of test cases are required to be implemented by the developers. For instance, the test case below evaluates whether the method returns 0 for a given argument with a 0 value:

%\vspace{3mm}

\begin{lstlisting}[language=Java, basicstyle=\small, basicstyle=\footnotesize\ttfamily]
@Test
public void MileToKmTest(){ 
	LengthConverter Converter = new LengthConverter();
    assertEquals(Converter.MileToKm(0),0,0.0001);}	
\end{lstlisting}

\vspace{3mm}

\fi

\subsubsection{Fine-Tuning the Model} 
Adopting code2vec architecture, we retrained the model with the cleaned dataset, including methods and corresponding unit tests. This process maps each source and test code fragment to a multidimensional space by generating an embedded vector for each sample. 

The model is re-trained with configuring the hyper-parameters as 9, 384, and 0.75, respectively, for the number of epochs, embedded vector size, and dropout keep rate. The training step finished after all 9 epochs.

\begin{table}
\scriptsize
\vspace{-5pt}
  \caption{The result of fine-tuning code2vec}
  \centering
  \vspace{-5pt}
 % \footnotesize
  \label{tab:fine_tuning}
 \begin{tabular}{ccccccccccc}
 \hline\noalign{\smallskip}
\textbf{}&\textbf{Precision}&\textbf{Recall}&\textbf{F1}\\
\hline\noalign{\smallskip}
 Pretrained & 63.1\% & 54.4\% & 58.4\% \\
 This work & 64.4\% & 52.1\% & 57.6\% \\
\hline
\end{tabular}
\vspace{-10pt}
\end{table}

Once the transformation of the training samples is completed, the model is ready to receive a new method as an input and generates the method's distributed embedded vector. The fine-tuned model, while in operation, maps the newly-generated embedding for the input in the same space, which it built during the training phase. The placement of the vector is in such a way that the methods with similar content be closer than those with different semantics (use of words) in syntactically-important paths (AST paths carrying the large weights of the method context). Then, as briefly mentioned below, the pipeline split into two different approaches. 
In the first approach, existing embedded methods with more than 90 percent similarity to the given method's embedding are identified, and finally returns, the test cases associated with the most similar methods, in the form of recommendations to the developer as the potential test cases. In the second approach, the test case with more than 70 percent similarity to any given method will be extracted and recommended to the developers. The reason behind the lower threshold in the second approach is that, generally, the methods and test cases have different structures and syntax.

\subsection{Measuring Similarity}

%We initially re-trained code2vec on 4,500 samples of methods and associated test case pairs.The re-trained model then generates the embedded vectors for the entire set of methods as well as test cases. The embedded vectors of both artifacts, methods, and test cases are then mapped to a high-dimensional space such that the closer vectors have more semantic and structural similarities with each other. All embedded methods and test cases will be used in the process of similarity calculation with any new method. Moreover, embedded test cases will be used as a bank of possible recommendations for any new method.
The assumption that this research is driven is that our fine-tuned model associates \textit{``similar''} methods in the source code to embeddings, which are placed closer to each other. Furthermore, in the same space, similar unit test cases are similarly transformed into closely-placed vectors. 
This section evaluates the assumptions we made.

For this, we initially assessed whether a correlation exists between the similarity of methods and their test cases. In other words, we investigated whether or not the methods with higher structural and semantic similarities tend to associate with a similar test pair as well.

\subsubsection{The Criterion of  Relevance}
Here we tested the assumption that methods with high semantic and structural similarity could have high similarity between their test cases. To measure the distance between embedded vectors, we measured the cosine similarity of the angle between the vectors. This measure is commonly used in the literature to calculate the similarity between the vectors in high dimensional space \cite{7577578}. We inferred that the same metric also serves our purpose because the cosine similarity does not depend on the magnitudes of the vectors but only on their angle. For instance, cosine similarity is $1$, $0$, and $-1$, respectively, for two proportional vectors, orthogonal vectors, and opposite vectors.

\subsubsection{Relevance of the Methods}

In this part of the study, we leverage the fine-tuned version of code2vec to generate embedding vectors. The fine-tuned model performance has reported in table \ref{tab:fine_tuning}. The model generates vectors with a fixed length, equal to 384, which is the size of the one before the last fully connected layer of the model.

Generating the embedded vectors for the entire methods and test cases, the cosine similarity between each existing pair of the embedded method vectors and the pairs of the embedded unit test vectors in the dataset were calculated.

The radar charts in Figure \ref{fig:use2} represent a subset of similar pairs. The circle radius displays a similarity measure between zero (closer to the circle center) and one (closer to the circle circumference line). In Figure \ref{fig:use11} the method pairs are represented in green, while orange is selected to illustrate the unit tests associated with the selected methods. As visible, the chart only displays the method and test pairs whose similarity measures between their methods were greater than 0.9 (90 percent similar). For each green point presenting a highly similar method pair, one relevant orange point represents the similarity score between their associated test cases.

Given the distribution of the orange test pairs in Figure \ref{fig:use11}, the majority of the test pairs' similarity scores are concentrated at the center of the radar, representing the pairs with a low similarity score (closer to zero). The rest of the scores are distributed further than the center with no identifiable pattern. This phenomenon is known as the \textit{explosion effect} lines, resembling the after-effect impacts of an explosion, potentially occurring in the center of the circle. In terms of visualization, this research aims to remove the explosion effect lines, achieving long, consistent yellow lines that reflect the pairs' similarity scores. Once achieved, one could observe the yellow to be more evenly within the circle area, as opposed to being mostly collected in the center and only a few lines reaching the perimeter of the circle.

\subsubsection{Relevance of Methods \& Test Pairs}

During this experiment, we observed the occurrence of a repetitive pattern for the majority of the similar method pairs. Here, we noticed that while the entire set of the highly-similar methods is not necessarily accompanied by test units that are also highly similar to each other, for a majority of these method pairs, there are several unit tests in the dataset which were highly similar to them. Associated test cases are also highly similar. For instance, in Table \ref{tab:sample} a method sample, labeled with 216, has the highest cosine similarity with the four listed methods. But the similarity between the test case of sample 216 with the test cases of these four methods is very different from each other. As such, we conducted another experiment in which we extracted the test cases of the entire methods which contained more than 90 percent similarity with a given method and further used the similarity between the method and test case as the second recommendation measure.

\begin{comment}
\begin{table}
\small
  \caption{Application 1: Method and Test Case similarity of sample \#216 with four other samples.}
  \centering
 % \footnotesize
  \label{tab:EEL}
 \begin{tabular}{ccccc}
% \hline\noalign{\smallskip}
 &$Measure_1$&$Measure_2$&&$Truth$\\\hline
\textbf{ID}&\textbf{CosSim}&\textbf{CosSim}&\textbf{LevSim}\mona{move to RQ2}&\textbf{CosSim}\\
 &($M_{216},M_{ID}$)& ($M_{216},T_{M_{ID}}$)&($M_{216},T_{M_{ID}}$)&($T_{216},T_{M_{ID}}$)\\
%\hline\noalign{\smallskip}
$M_{1}$ (\#2,848) & 1.0 & -0.067 & 430 & 0.189  \\
$M_{2}$ (\#3,081) & 1.0 & \textbf{0.029} & \textbf{126} & \textbf{0.802}  \\
$M_{3}$ (\#1,668) & 1.0 & 0.090 & 445 & 0.542  \\
$M_{4}$ (\#2,233) & 1.0 & \textbf{0.107} & \textbf{149} & \textbf{0.990}  \\

\hline
\end{tabular}
\end{table}
\end{comment}

 Given the previous observation, we assessed whether or not an appropriate test case is semantically similar to its corresponding method. A positive answer to this question allows leveraging the aforementioned similarity score to find more useful recommendations for the developers.

Two methods \textit{``some''} semantic similarity do not necessarily share a similar context. In such cases, therefore, the test cases required for assessing the functionality of the two methods may as well appear to be different.

\begin{table}[htb!]
\scriptsize
  \caption{An example of a random sample ($M_{216}$)'s similarity to methods and test cases present in the dataset.}
  \centering
 % \footnotesize
  \label{tab:sample}
 \begin{tabular}{ccc}
 \hline{}
 %\cline{2-3}\noalign{\smallskip}
                                             & \textbf{Methods Similarity}   & \textbf{Test Cases Similarity} \\
                                             & \textbf{$(M_{216},M_{ID}$)}   & \textbf{($T_{216},T_{M_{ID}})$}\\
\hline\noalign{\smallskip}
{$M_{1}$}                  & 1.0                           & 0.189                          \\
\cellcolor{Gray}{$M_{2}$}  & \cellcolor{Gray}1.0           & \cellcolor{Gray}0.802          \\
{$M_{3}$}                  & 1.0                           & 0.542                          \\
\cellcolor{Gray}{$M_{4}$}  & \cellcolor{Gray}1.0           & \cellcolor{Gray}0.990          \\\hline
\end{tabular}
\vspace{-10pt}
\end{table}

\subsection{Recommendation Criteria}
 With respect to the two alternative focus of the term \textit{similar} in the hypothesis, we developed two approaches in which the criteria for recommending unit tests differ. The first application prioritizes the intended functionality of a given method to make recommendations, while the second emphasizes more the similarity of the features which are under test to select a unit test for a recommendation. We then evaluated both variants to assess the validity of the assumption made earlier. 

\subsubsection{First Approach (Functionality-Oriented)}
The first approach, in which similar functionality of the method pairs is the focus of the recommendation, holds an underlying assumption. The hypothesis to be tested here assumes the more `similar' the methods in terms of functionality, the more similar their unit tests are as well. As such, for a given method, the model recommends associating unit tests with the methods whose embedded vectors are closely located to the input method's embedded vector.
 
To implement the first approach, for the vectorized representation of a given method, first, the closest embeddings are marked, if and only if they are associated with a method in the source code, not a test unit in a test suite. Second, among the most similar methods, the priority is the unit test, in which embedding is closer to the input embedding. This means this variant favors similarity in the functional purposes of the input method over the feature which is being tested.

\subsubsection{Second Approach (Structure-Oriented)} Here the method's primary functionality is majorly ignored, meaning that the priority of selecting a unit test recommendation is independent of the functionality, and instead, depends on the feature I want to test. If I am testing the same feature (such as input and output of the method) whatever the method does (print, calculates, measure, etc) regardless of the functionality of the given method, the unit tests are more similar.

The development of the two application variants aims to assess whether unit tests closely relate to the method's functionality (often a common assumption) or to the feature which is tested. For instance, the two unit tests, such as maxPayement() and minPayment(), intended to test the two scenarios of over-payment and under-payment, tend to be more similar to other unit tests which are relevant to the Payment() functionality, such as cashPayment(), or to unit tests which test the maximum and minimum capacity of other elements, such as minArray() and maxArray().

\begin{figure}
     \centering
     \begin{subfigure}[b]{0.54\columnwidth}
         \centering
         \includegraphics[width=\columnwidth]{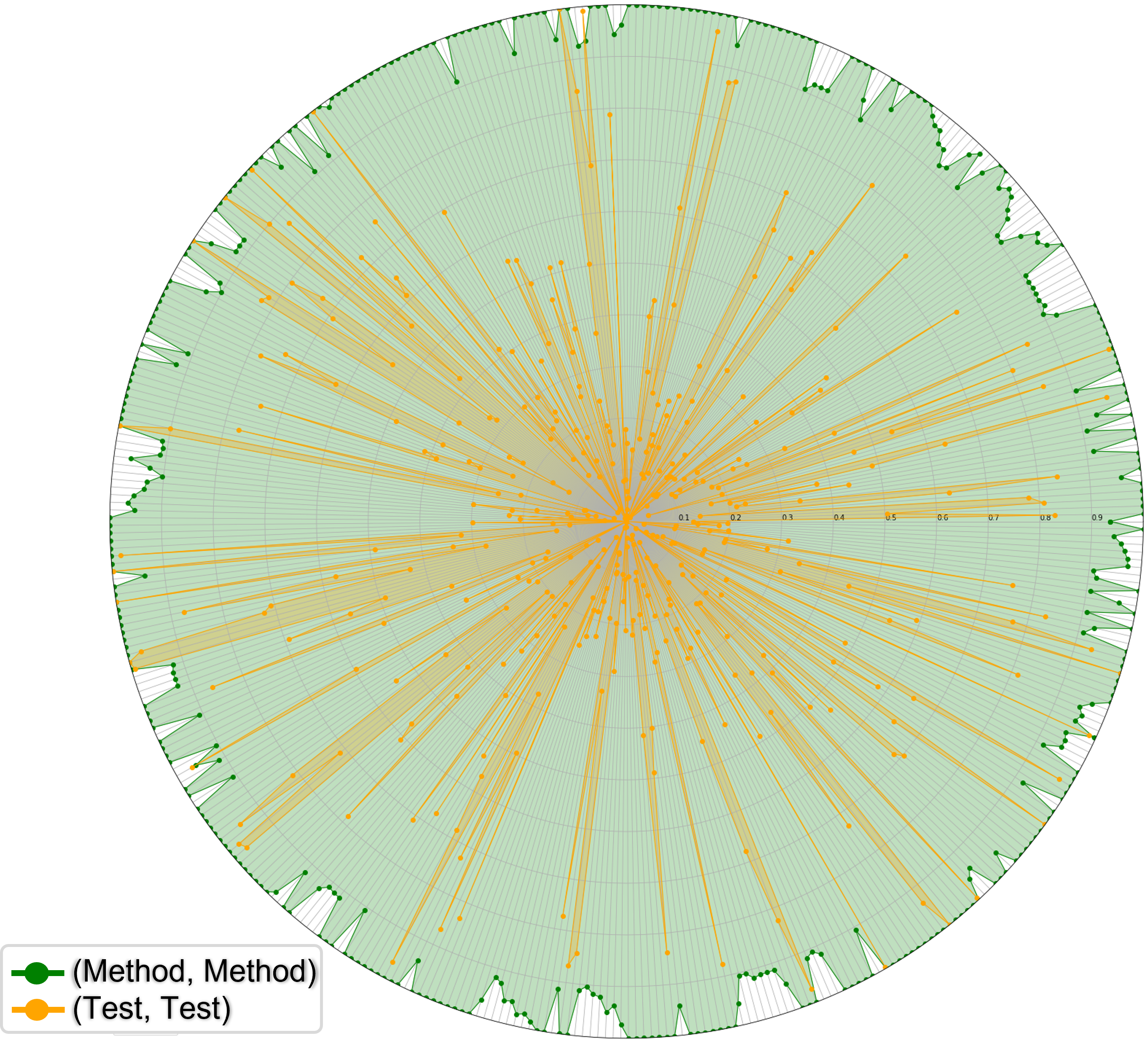}
         \caption{All Sample}
         \label{fig:use11}
     \end{subfigure}
     \hfill
     \begin{subfigure}[b]{0.45\columnwidth}
         \centering
         \includegraphics[width=\columnwidth]{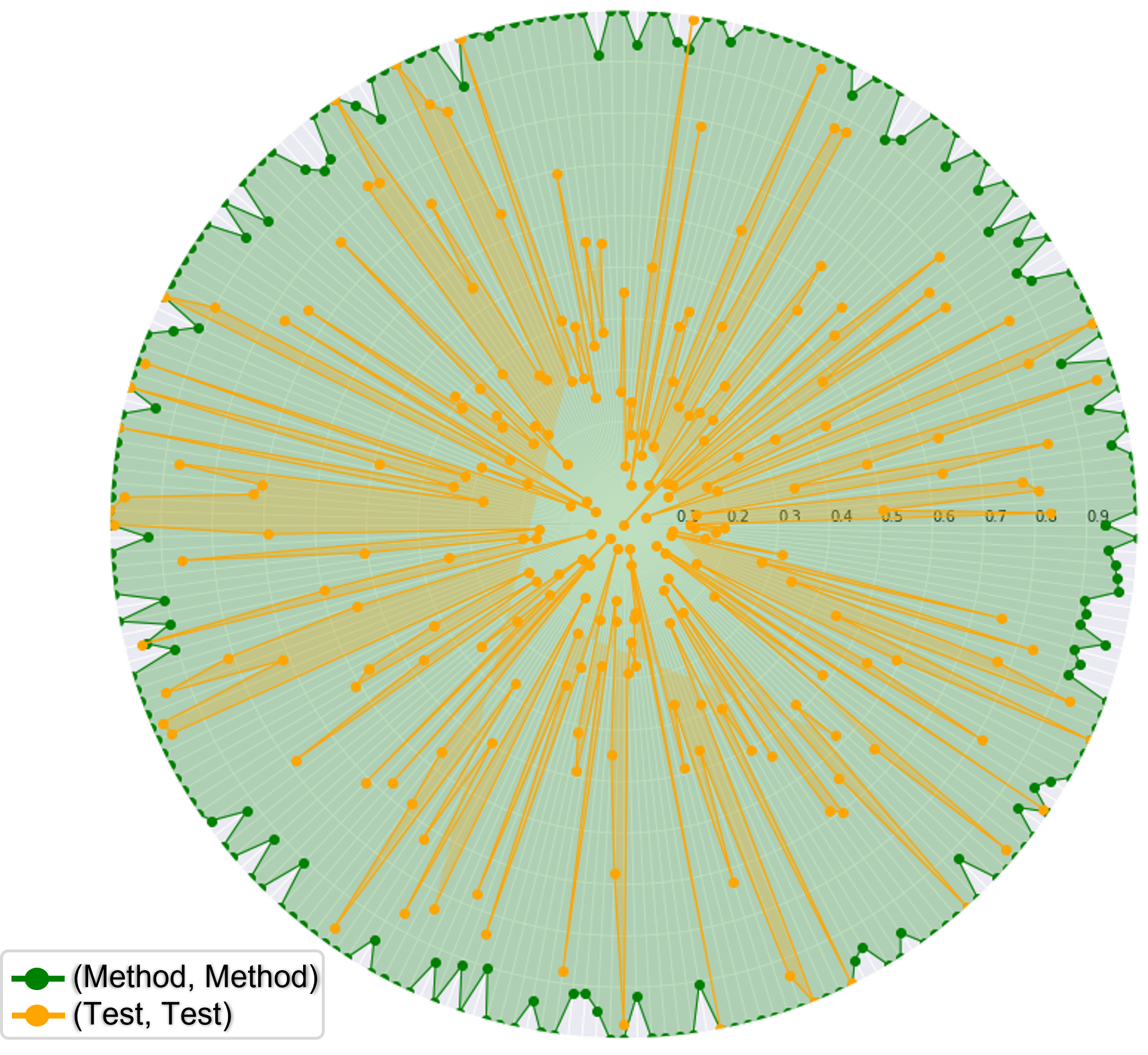}
         \caption{First Approach}
         \label{fig:use22}
     \end{subfigure}
     \hfill
               \begin{subfigure}[b]{0.45\columnwidth}
         \centering
         \includegraphics[width=\columnwidth]{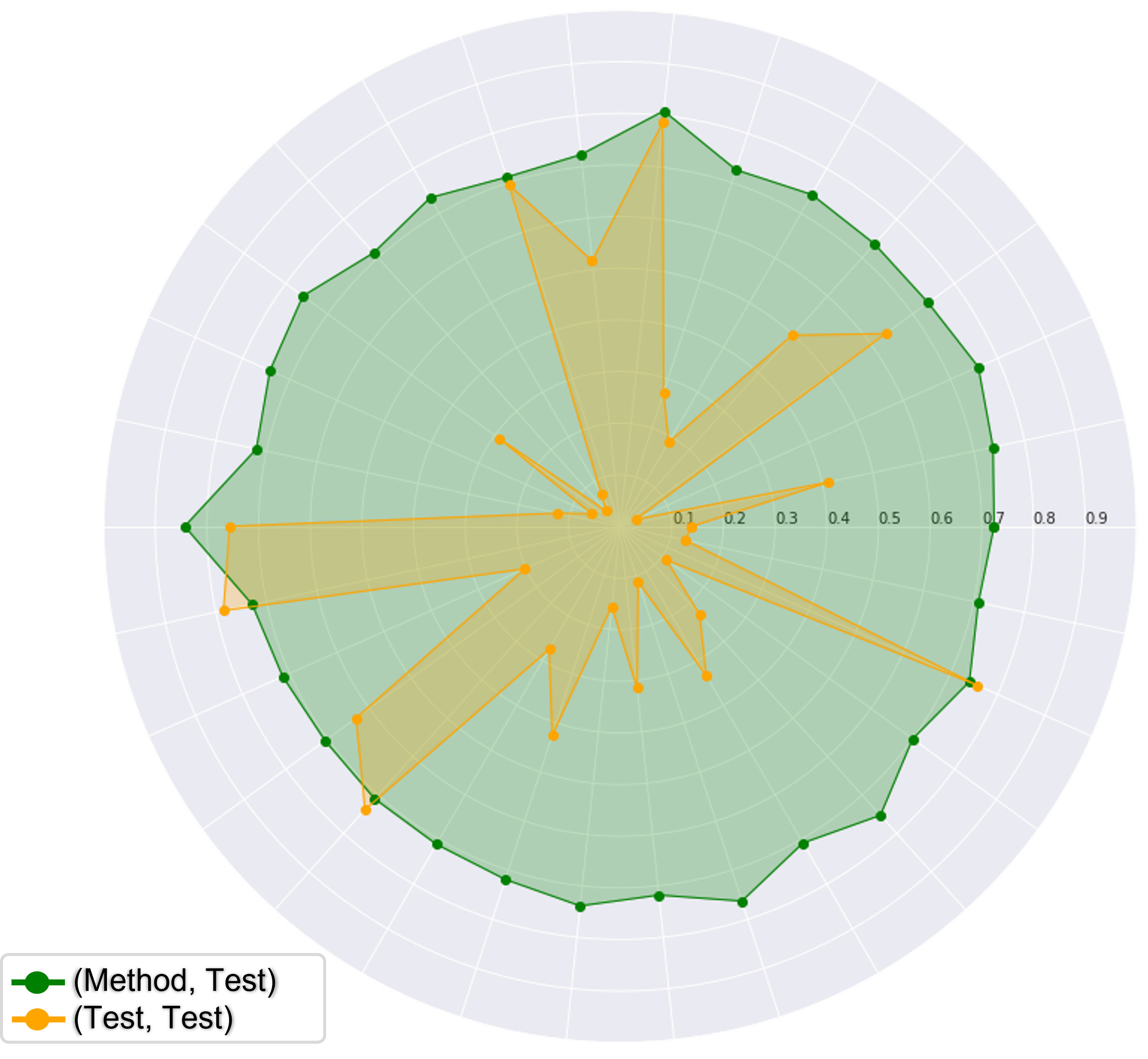}
         \caption{Second Approach}
         \label{fig:use33}
     \end{subfigure}
     \caption{Radar Charts}
        \label{fig:use2}
\end{figure}

\begin{figure}[ht]
\vspace{-5pt}
\centerline{\includegraphics[width=.6\linewidth]{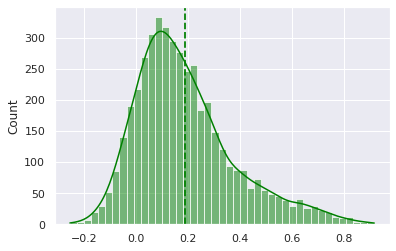}}
\caption{Histogram of cosine similarity between Methods and their Test Cases.}
\label{fig:use3}
\vspace{-10pt}
\end{figure}
\section{Evaluation}
\label{evaluation}

%\vspace{3mm}

This section evaluates the proposed applications in order to answer the research questions below:

\subsection{RQ 1: Given the structural and semantic similarities between methods, how similar are their associated test cases?}

To answer this research question, we computed the distance between each recommended unit test while preserving the structural and semantical properties of the code and the developer-written unit test as the ground truth. The distance is captured through computing the cosine similarity between the associated embedded vectors of recommendations and actual unit tests.   The results are reported in Table \ref{tab:mean}. As shown, the mean of the cosine similarity between the oracle (\begin{math}{T_{a}}\end{math}) and recommended test cases (\begin{math}{T_{b}}\end{math}) in the first approach is 0.48, which is larger than the same similarity measure in the second approach, which is equal to 0.35. A significance test, namely the student t-test, with a p-value of 0.05, verified the significance of the difference between the means of the two populations.
The difference can be visually inferred by comparing Figures \ref{fig:use11} and \ref{fig:use22} with Figure \ref{fig:use33}.

\begin{table}[htb!]
\scriptsize
%\vspace{-5pt}
  \caption{Frequency percentages of 50\% \& 70\% similarity between samples, recommended units by approaches 1 \& 2.}\vspace{-5pt}
  \centering
 % \footnotesize
  \label{tab:TTPercentage}
 \begin{tabular}{cc||cc}
 \hline%\noalign{\smallskip}
\textbf{Similarity}&\textbf{All Samples}&\textbf{Approach 1}&\textbf{Approach 2}\\
\hline%\noalign{\smallskip}
 lower than 50\% & 69\% & 56\%& 70\% \\
 lower than 70\% & 80\% & 69\% & 83\% \\
\hline
\end{tabular}
\vspace{-5pt}
\end{table}

Furthermore, as shown in Table \ref{tab:TTPercentage} for the first approach, 13\% and 11\% of the samples are further included within the samples with similarities of more than 50\% and 70\%, respectively.  As shown in Figure \ref{fig:use22}, an increase in the average similarity leads to a decrease in the explosion effect and to smoothened edges. On the other side, the second approach does not show any significant improvement. Please note that the low number of recommended test cases by the second approach was expected due to the low similarity between the methods and test cases as shown in Figure \ref{fig:use3}. The comparison of the three Figures \ref{fig:use11}, \ref{fig:use22} and \ref{fig:use33} illustrates that the explosion effect is reduced, and  the lines are more uniformly distributed from Figure \ref{fig:use11} to \ref{fig:use22}, and similarly from Figure \ref{fig:use11} to \ref{fig:use33}, meaning that more similar test cases are extracted in both approaches, while this pattern is more successfully consistent in the first approach, in Figure \ref{fig:use22}, in comparison to the second approach in Figure \ref{fig:use33}.

\begin{table}[htb!]
\scriptsize
\vspace{-5pt}
  \caption{Similarity between (Method, Method), (Method, Test), and (Test, Test) pairs}
  \centering
  \vspace{-5pt}
 % \footnotesize
  \label{tab:mean}
 \begin{tabular}{ccc||cccc}
 \hline
 %\cline{33-7}\noalign{\smallskip}
  & \multicolumn{2}{c}{\textbf{All Samples}}  & \multicolumn{4}{c}{\textbf{Recommendations}}\\
  & \multicolumn{2}{c}{\textbf{Cosine}}       & \multicolumn{2}{c}{\textbf{Approach 1}}         & \multicolumn{2}{c}{\textbf{Approach 2}}\\\hline
  \tiny 
  & \tiny${M_{a},M_{b}}$ 
  & \tiny${T_{a},T_{b}}$ 
  & \tiny${M_{a},M_{b}}$ 
  & \tiny${T_{a},T_{b}}$ 
  & \tiny${M_{a},T_{b}}$ 
  & \tiny${T_{a},T_{b}}$\\\hline

%\hline\noalign{\smallskip}
\multicolumn{1}{c}{\footnotesize\textbf{Count}} &  421  &  421  &  232  &   232  &  30   &  30   \\
\multicolumn{1}{c}{\cellcolor{Gray}\footnotesize\textbf{Mean}}  &  \cellcolor{Gray}0.97 &  \cellcolor{Gray}0.37 &  \cellcolor{Gray}0.98 &   \cellcolor{Gray}0.48 &  \cellcolor{Gray}0.73 &   \cellcolor{Gray}0.35 \\
\multicolumn{1}{c}{\footnotesize\textbf{Std.}}  &  0.03 &  0.29 &  0.03 &   0.30 &  0.03 &   0.25 \\
\multicolumn{1}{c}{\cellcolor{Gray}\footnotesize\textbf{Max.}}  &  \cellcolor{Gray}1.0  &  \cellcolor{Gray}1.0  &  \cellcolor{Gray}1.0  &   \cellcolor{Gray}1.0  &  \cellcolor{Gray}0.84 &   \cellcolor{Gray}0.78 \\
\multicolumn{1}{c}{\footnotesize\textbf{Min.}}  &  0.9  &  0    &  0.9  &   -.08 &  0.7  &   0.03 \\
\hline
\end{tabular}
\vspace{-5pt}
\end{table}

The scenario in which the evaluation results revealed that Approach 1 performs more accurately in proposing a required unit test would show that the required unit tests are often more correlated with the method functionality. For instance, in the given payment() example, the unit tests which are required are more similar to another payment function's unit tests. In the reverse scenario, the lesson learned relative to the given dataset is that the unit tests are more semantically and structurally similar to the min and max functions unit tests. 

Figure \ref{fig:use22}, represents the similarity scores of the proposed test cases by our first approach for the same method pairs in Figure \ref{fig:use11}. This new radar chart is, therefore, the updated representation of the test pair similarity between the same method points after our proposed approach selected a similar test case. As shown, the similarity between the test cases which our method selected as the recommendation is much higher than the general similarity between all the test pairs. 

Furthermore, Figure \ref{fig:use33} shows that more than 70 percent similarity between any given method and the test cases obviously increases the chance of recommending much more related test cases.
Analyzing the radar charts in Figure \ref{fig:use2} shows that the explosion effect is decreased, and a large number of the yellow points, representing the similarity of the pairs, are relocated farther from the center and closer to the perimeter of the circle. 

As reported in Table \ref{tab:TTPercentage}, after applying the first and second approaches, once similar pairs, with similarity above 50\%, and once again, pairs with similarity above 70\% were selected. The extraction of similar pairs was repeated three times, first from the entire 4,500 embedded samples (second column), the recommended unit tests by the first (third column) and second (last column) approaches. The statistics of the entire sample are given for comparison purposes. 
 
As shown, 69\% and 80\% of the entire samples, in general, have less than 50 and 70 percent similarity to the expected targets, respectively. 
The selection of test cases in the first approach, among all extracted samples, leads to 56\% average similarity below 50\% and 69\% below 70\% similarity between recommended test cases and expected test cases.
 
Although the second approach only slightly changes the distribution of samples, the effect on the distribution of the recommended test cases is significantly shifted by the first approach. The first approach decreases the frequency of samples with less similarity with the targets, from 69\% and 80\% to 56\% and 69\% respectively. In other words, the proposed approach decreases the  explosion effect and distributes the similarity lines more uniformly, which means the frequency of the less similar test cases between the given method and the tests are decreased.
Both approaches' results with more details are reported in Tables \ref{tab:mean}.

\subsection{RQ 2: Recommending the associated test cases for the methods with syntactical and semantic similarity, how useful are these recommendations in reducing developers’ effort?}
We estimated the developers' efforts through measuring the edit distance between the recommended unit test and the actual unit present in the dataset. 

Edit distance counts the minimum number of token-level operations (insertion, deletion, and substitution) required to transform one of the two given strings to the other one. The quantification of the edit distance between two strings is broadly applied in the majority of domains. For instance, in NLP, the edit distances are used to search and identify the most similar words, in a dictionary, to a typed word for auto-correction purposes. As another example, in bioinformatics, the similarity between the DNA strings is computed to search for similar DNAs.

To measure and compare the usability of the two approaches proposed in this research, we computed the token-based edit distance, the Levenshtein distance \cite{levenshtein1966binary}, between the recommended and the existing test cases. Levenshtein distance is commonly used for determining the similarity between two strings or source codes by counting the number of deletions, insertions, or substitutions required to transform one string to another. For this reason, we relatively estimated the usability of each proposed approach in reducing the developers' effort in developing unit tests.

Tables \ref{tab:lev_eval} shows the first approach leads to recommendations, which distance to the unit test written by a developer is less than the second approach. Please note that a lower distance represents less tokens need to be edited to generate the answer from the recommended test case. Therefore the approach with a lower distance represents the more useful method in recommending unit tests. As an example, a recommended test case for method 216 has been shown below.

\begin{table}
\scriptsize
\vspace{-5pt}
  \caption{Levenshtein distance between oracles and Recommended test cases for both approaches.}
  \centering
 % \footnotesize
  \label{tab:lev_eval}
 \begin{tabular}{ccc||cc}
 \hline
 %\cline{2-5}\noalign{\smallskip}
 &\multicolumn{2}{c||}{\textbf{Approach 1}}&\multicolumn{2}{c}{\textbf{Approach 2}}\\\hline
 &${Lev.}$&
 ${Length}$& 
 ${Lev.}$& 
 ${Length}$\\\hline
%\hline\noalign{\smallskipskip}
\multicolumn{1}{c}{\textbf{Count}} & 232    & 232    & 30     & 30   \\
\multicolumn{1}{c}{\cellcolor{Gray}\textbf{Mean}}  & \cellcolor{Gray}301  & \cellcolor{Gray}439    & \cellcolor{Gray}706    & \cellcolor{Gray}757  \\
\multicolumn{1}{c}{\textbf{Std.}}  & 306    & 372    & 612    & 636  \\
\multicolumn{1}{c}{\cellcolor{Gray}\textbf{Max.}}  & \cellcolor{Gray}1,708  & \cellcolor{Gray}2,292  & \cellcolor{Gray}2,312  & \cellcolor{Gray}2,521 \\
\multicolumn{1}{c}{\textbf{Min.}}  & 0      & 64     & 111    & 79   \\
\hline
\end{tabular}
\vspace{-10pt}
\end{table}

\vspace{1mm}

\noindent \textbf{\small Oracle:}

\lstdefinestyle{mystyle}{
tabsize=3, 
breaklines=true,
showstringspaces=false,
numbersep=8pt}
\lstset{style=mystyle}

\begin{lstlisting}[language=Java, basicstyle=\small, basicstyle=\footnotesize\ttfamily]
@Test(expected = StackTooSmallException.class) 
public void testSSTORE_3() { 
    program = getProgram("602255"); 
    try { 
        vm.step(program); 
        vm.step(program); } 
    finally { assertTrue(program.isStopped()); } }
 
\end{lstlisting}

%\vspace{3mm}

\noindent \textbf{\small Recommendation:}

\begin{lstlisting}[language=Java, basicstyle=\small, basicstyle=\footnotesize\ttfamily]
@Test(expected = StackTooSmallException.class) 
public void testDUPN_2() { 
    program = getProgram("80"); 
    try { vm.step(program); } 
    finally { assertTrue(program.isStopped()); } }

\end{lstlisting}

%\vspace{4mm}
\section{Threats to Validity}
\label{Validity}

As discussed earlier, the embedding algorithm used in the research, code2vec, was unable to generate the embedded vectors of the entire source code fragments. Hence, several code fragments had to be removed from the study due to this limitation, which may have impacted our evaluation results. Moreover, since an increase in the number of samples may lead to an increase in the chance of finding more similar methods and unit tests, the lack of enough samples is another threat to the evaluation results. 

Further, although the samples (methods) had oracles (expected test cases), there is no guarantee that the oracles are the only test cases that the developers require. The required test cases depend on several factors, such as the functionality of the method, the developers' purpose of testing, and the class that the method belongs to. Hence, in several cases, the recommended test cases or even the oracle were not what the developers required. However, the proposed approach tends to perform as a recommendation system, which suggests similar code snippets to the developers with the purpose of decreasing the effort which is necessary in writing test cases.
\section{Conclusion \& Future Work}
\label{conclusion}

Maintaining test suites is tedious and time-consuming for developers. To this end, a test case recommender has been proposed, which provides the closest unit test among those present in a test suite to what the model recognizes is the developer's need, which in turn reduces the effort of writing a unit test from scratch. For this, initially, the entire methods and unit tests are embedded into a continuous vectorized representation by a neural network model, known as code2vec, which preserves the structural and semantic characteristics of the artifacts.  According to the cosine similarity between the generated vectors two approaches were implemented to select the closest vector for recommendation. One prioritizes the structure and semantic similarity between the methods, while the other one favors the similarity between the methods and test cases. 
Finally, the results show that based on the similarity between the embedded vectors, relevant test cases can be extracted. Also, the Levenshtein distance shows that the proposed approach decreases the number of changes that developers need to apply to recommended test cases based on their purposes. As future work, we tend to include more ``known'' features about the method under the test to search for similar unit tests, as well as to extend the search space of recommendations to online information.

%\vspace{11mm}

\bibliographystyle{IEEEtran}
\bibliography{sample-base}

\end{document}